# Large Language Models and Arabic Content: A Review

**Haneh Rhel**

University of Strathclyde

16 Richmond Street, Glasgow G1 1XQ

haneh.rhel@strath.ac.uk

**Dmitri Roussinov**

University of Strathclyde

16 Richmond Street, Glasgow G1 1XQ

dmitri.roussinov@strath.ac.uk

**Abstract**

Over the past three years, the rapid advancement of Large Language Models (LLMs) has had a profound impact on multiple areas of Artificial Intelligence (AI), particularly in Natural Language Processing (NLP) across diverse languages, including Arabic. Although Arabic is considered one of the most widely spoken languages across 27 countries in the Arabic world and used as a second language in some other non-Arabic countries as well, there is still a scarcity of Arabic resources, datasets, and tools. Arabic NLP tasks face various challenges due to the complexities of the Arabic language, including its rich morphology, intricate structure, and diverse writing standards, among other factors. Researchers have been actively addressing these challenges, demonstrating that pre-trained Large Language Models (LLMs) trained on multilingual corpora achieve significant success in various Arabic NLP tasks. This study provides an overview of using large language models (LLMs) for the Arabic language, highlighting early pre-trained Arabic Language models across various NLP applications and their ability to handle diverse Arabic content tasks and dialects. It also provides an overview of how techniques like fine-tuning and prompt engineering can enhance the performance of these models. Additionally, the study summarizes common Arabic benchmarks and datasets while presenting our observations on the persistent upward trend in the adoption of LLMs.

**Keywords:** Large Language Models, LLMs, Arabic language, Natural Language Processing, NLP tasks.

## 1. Introduction

The development of large language models (LLMs) has significantly changed the natural language processing (NLP) field in recent years [1] [2] [3]. Various corpora containing billions of words have been used to pre-train these models, enabling LLMs to deal with different tasks, including processing semantics, translation, dialects, linguistic structures, and languages. Despite these advancements, challenges related to the coverage of diverse languages still need to be addressed. Arabic is a challenging language that has attracted increased research attention in addressing specific language-processing tasks over recent years [4]. It presents unique opportunities and difficulties for large language models due to its *dialect diversity*, *complex grammar*, *rich morphology*, and others [5] [6][7]

Various studies have highlighted the limitations of utilizing LLMs with Arabic content. According to [8], the development and implementation of efficient LLMs face challenges due to Arabic's rich morphology, the scarcity of high-quality data and benchmark datasets, as well as complexities in semantics, emotion recognition, text simplification, and complex word detection. As a result, researchers have been working [9] to address such difficulties by adopting innovative approaches in training procedures, model design, and data collection.

On the other hand, there are still various advantages of using LLMs with Arabic tasks. For instance, several experiments have presented superior results in the following tasks: text generation, summarization, grammatical error correction [5], Arabic AI detection [10], and question-answering. LLMs could probably help bridge the gap between Modern Standard Arabic (MSA) and Classical Arabic (CA), as well as regional dialects [11], assisting in enhancing natural and contextual communication.

Starting from GPT-3, the Generative Pre-trained Transformer (GPT) line of LLMs has been one of the best-performing during the last 3 years on standard benchmarks, indicating outstanding proficiency in natural language processing tasks[2]. It works by processing the user's commands/prompts based on which corpus is pre-trained.

ChatGPT [12] as a Conversational GPT model was the first LLM that achieved significant success in text generation; it was then followed by Bard and Claude models.[13]. Therefore, in recent years, millions of users around the world have used these intelligent models in both research and industrial venues [4]. These Conversational models and others, typically known as chatbots, have become more essential to our routine lives. They work as a personal advisor on our smartphones, engaging frequently with humans using different formats such as voice/text, utilizing natural language [6].

In addition to OpenAI, companies such as Google (Palm, Bard, Gemini)[14], Meta (Llama)[15], Anthropic (Claude)[16], Microsoft (Copilot)[17], among others, and have continuously improved their models since the first model's version in various directions by enhancing the following points: data quality, dataset size, training tools/techniques, and number of parameters to get pure data and achieve multiple tasks in a wide range of domains and languages. For instance, the size of the GPT family model has increased significantly over recent years. It expanded almost 8,500 times, starting from ChatGPT to the most advanced models published recently in 2024 as GPT-4o/GPT-4o mini and o1 preview /o1 mini.

In general, the incorporation of LLMs promises to open up new possibilities for language classification, generation, and interaction in the Arabic-speaking world as the area of Arabic language AI expands. Through the utilisation of these sophisticated language models, we may discover novel techniques and possibilities for enhanced communication, text complexity detection, content simplification, knowledge exchange, and cross-cultural interaction. Although using LLMs has received more attention recently, most studies have been conducted in English, as illustrated in Figure 1.

This paper provides an overview of large language model (LLMs) applications to Arabic contexts and reviews related studies focusing on recent large language models such as ChatGPT, particularly GPT-3 and subsequent versions. It also includes discussing pre-trained Language Models (PLMs) like AraBERT, MARBERT, and QARIB, outlining their historical development, limitations of Arabic publication, Language-specific challenges, common Arabic benchmarks and datasets, observed trends, and future directions.

## 2. Arabic LLM Publications

Arabic is one of the most widely used languages on the Internet [18] . It is the official language of twenty-seven countries and is spoken by approximately 422 million people in the Arabic world [9]. As depicted in Figure 1 illustrates that few studies have focused on large language models in Arabic so far. In contrast, English is the primary language of conversational (AI) model research.

*Figure 1(a) highlights a notable increase in the published Arabic Language Models (PLMs and LLMs) articles at two of the leading annual conferences, ACL and EMNLP. Similarly, Figure 1(b) shows a sharp rise in the published English Language Models studies during the period from 2019 to 2024.*

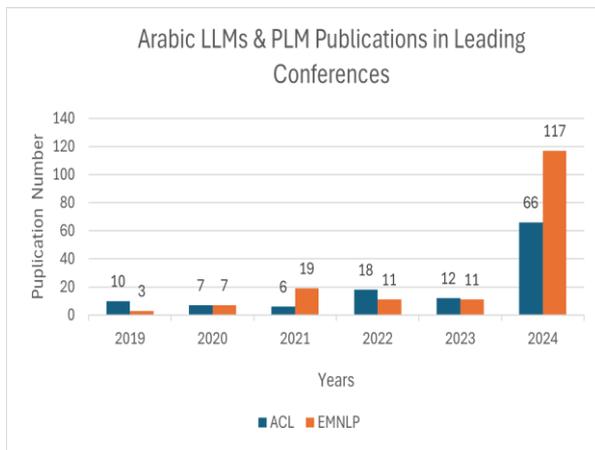
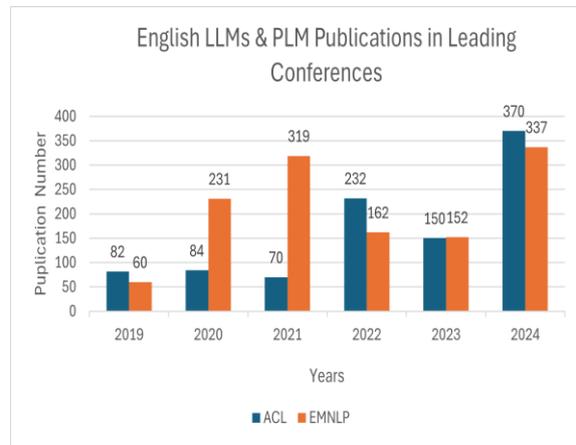

**Fig. 1.** (a) Arabic LLM & PLM papers     **Fig. 1.** (b) English PLM &LLM papers

The limited number of studies related to the LLMs in the Arabic context is likely due to the complex features of the Arabic language, such as rich morphology, orthography, lexical ambiguity, diacritics, multi-Arabic dialects, and right-to-left script, as detailed by Boudad [9] These factors make Arabic particularly challenging language for students and researchers.

To explicitly discuss Arabic LLM studies that reflect several of these features and frequently use terms such as CA, MSA, and DA, the next paragraph briefly defines these terms and highlights their role in the complexity of the Arabic language.

## 3. Arabic Language Specifics

There are three categories of the Arabic language:

(1) Classical Arabic (CA), mostly found in religious and historical writings, but still generally understood because it appears in the Holy Quran [19]. (2) Current or modern standard Arabic (MSA), a modern and simple form of CA, is the most understood language among Arabic speakers and commonly utilized in formal communication, journalism, and education, and everywhere in the different countries that speak Arabic [18]. (3) Dialectal/colloquial Arabic (DA), where each society has its own dialect. There are approximately 30 significant

Arabic dialects that are used daily by Arabic speakers rather than MSA, which show phonological, morphological, and lexical differences compared to MSA [20] .

**Rich Morphology**:   Arabic Morphological content has a root-and-pattern structure, similar to other Semitic languages. Each Arabic word should have a root, which is the basic meaning of the word, and typically consists of consonants. Vowels or non-consonant letters can be added as a special pattern to produce numerous linked words.

 For example, the word "كتب" is a root word with three main letters "ك.ت.ب" = "k,t,b" means "write" and is pronounced as "Kataba", if we add a consonant to become "كتاب" that is pronounced, "ketab", means "book". Arabic words can provide multiple morphological aspects, as follows: derivation, inflection, and agglutination, which are introduced in more detail by [9] and  [10]. In the same manner, Japanese, Hungarian, Turkish, Swahili (Niger and Congo Language), and some other languages, have a complex morphological structure that leads to create a wide multiword form of one single root word.

**Orthography**: In contrast to Latin languages, Arabic does not include upper/lower-case letters. It uses twenty-eight consonant letters and only three vowels. However, the Arabic script also employs **diacritical** markings as short vowels. These are positioned either above or below the letters to indicate how each word should be pronounced and to make its meaning clearer. Likewise, Kurdish (Sorani), Urdu, Persian (Farsi) and Hebrew languages also do not use uppercase or lowercase letters, and the diacritical marks in these languages are used for pronunciation and to differentiate between words' meanings.

**Diacritical**: Unlike proficient Arabic speakers, Arabic learners (non-proficient Arabic speakers) or children's books used diacritical marks.  For instance, in Table1**,** the word "علم" consists of three letters "ع","ل","م" = "A","l", and "m" in Arabic without diacritics can provide more than one meaning. It gives one meaning when adding different positioned diacritic marks as follows:

Table1. One word with different positioned diacritic marks presents different meanings

| Word with diacritics | Pronounce | Meaning |
|---|---|---|
| عَلَّمَ | Allama | Teach |
| عَلَمَ | Alam | Flog |
| عَلِمَ | Alema | Informed |
| عِلْم | Elem | Knowledge |

**Ambiguity**: Almost any content with no diacritical marks can probably present a lexical ambiguity problem, which could affect computational systems as well as Arabic learners.

## 4. Review of Related Works

### 4.1 Pre-trained Arabic Language Models

Several pre-trained Arabic language models have been developed to serve different purposes. However, this paper focuses on the most common models in the NLP field.

The unique features of the Arabic language, along with the limited data resources, pose significant challenges for researchers and developers in processing Arabic text. To address these obstacles, various studies have been conducted to develop, evaluate, understand, and enhance the efficiency of Pre-trained language models. This section presents an overview of some of these studies, which selected starting with a seed subset of papers found by Google Scholar using the query "pre-trained language models Arabic", and then we proceeded to explore the papers mentioned in this set of papers and the papers that cite them, and we tried to bring up PLMs research on various NLP applications.

A study conducted by [21] focused on AraBERT, which was particularly developed to address the unique features and challenges of the Arabic and languages other than English. AraBERT was adapted from BERT, a commonly used pre-trained language model that achieved superior success in several NLP tasks for English. [21] evaluated the effectiveness of the AraBERT model on a range of Arabic language understanding tasks such as Sentiment Analysis (SA), named entity recognition (NTR), and Question Answering (QA). It was pre-trained on a large Arabic dataset, including 70M sentences from different Arabic open-source corpora and news websites. Masked language modeling (MLM) and Next Sentence Prediction (NSP) were used as training objectives. A Sub-Word Tokenisation method was used for data processing, employing Farasa segmentation (a sub-word tokenization method) to divide words into roots, prefixes, and suffixes, which helped reduce redundancy and managed the complexity of the Arabic lexical. AraBERT showed state-of-the-art performance on most of the above-mentioned Arabic NLP tasks and demonstrated a strong foundation for further studies. However, only a limited number of Arabic dialects were included in the trained dataset.

Google released the Arabic BERT model in 2019 as part of mBERT (multilingual BERT) models, which addressed Arabic and other than English languages. The Arabic BERT was conducted properly on general Arabic tasks but with less performance with special dialects that need more understanding of the Arabic language morphology, whereas AraBERT (monolingual language) was not released only for Arabic and developed with adjusting the characteristics of the Arabic language. It was built exclusively for Arabic (monolingual language), trained on large Arabic datasets including different Arabic corpus, and used Farasa segmentation as its unique approach.

The Arabic-BERT model was developed by the Qatar Computing Research Institute (QCRI) and the Massachusetts Institute of Technology (MIT). It was released as a pre-trained Arabic model that trained on the OSCAR corpus, which is a large Arabic dataset, including articles, Arabic books, both MSA and CA literature, and websites, also using Arabic Wikipedia datasets that contain approximately 1.5 million articles to provide a comprehensive representation of the language. It achieved a strong performance in Arabic NLP tasks. It used a multi-layer transformer, which is an encoder, the same as the common BERT architecture, which was originally trained on a large English corpus. The Arabic-BERT used a collection of Arabic characters and diacritics, which were about 30,000 subworlds with a model size of 110M parameters. Masked Language Modeling (MLM) was used and trained to predict tokens (words) in the input sentences. In addition to that, the Next Sentence Prediction (NSP) model predicted which sentence is relative to the original text.

Undoubtedly, the distinctions between AraBERT and Arabic-BERT in terms of pretrained corpus, data size, tokenization techniques, and architectural design have significantly advanced the Arabic natural language processing (NLP) field. Both models are well-suited for modern standard Arabic (MSA) and have demonstrated robust performance across various Arabic NLP tasks, as demonstrated by [22] Arabic-BERT has been effectively applied to various tasks, including sentiment analysis, question answering, text classification, named entity recognition, and part-of-speech tagging. Its success paved the way for researchers to further develop modern Arabic model, such as MARBERT (2020) which employed the mask language modeling (MLM) strategy, although highly computational, it improved learning contextual representation, as well as the AraELECTRA (2021) model, which enhanced the model's performance with fewer training resources using the more efficient strategy called replaced token detection(RTD). etc. In contrast, recent LLMs utilised advanced strategies like Reinforcement Learning from Human Feedback (RLHF), which required massive computational resources but with more data efficiency.

In 2020, the **QARIB model** was also released by the Qatar Institute (QCRI) with a larger size of 165MB and 50,000 subworlds and showed better performance [23]. Despite achieving significant results, both Arabic-BERT and QARIB models have important limitations that have been highlighted in several studies, such as requiring computational resources and significant computer memory and having limited domain knowledge resulting in the general training of Arabic datasets, which might not cover all domains. Also, there was an overfitting risk and a lack of interpretability.

**MARBERT** is one of the first powerful Arabic models developed in 2020 by Mohammed bin Salman College of Business and Entrepreneurship (MBSC) in collaboration with the University of Edinburgh for Arabic NLP tasks. It showed state-of-the-art results, particularly in dialectal Arabic, and was also utilized to support Modern Standard Arabic MSA. As indicated above, MARBERT is utilised in various Arabic NLP tasks such as machine translation tasks with "Arabic-English translation tasks", the analysis of Arabic text sentiment, text classification tasks "Labelling Arabic texts into different genres or topics", and conversational Arabic AI model that can understand the Arabic input and respond with Arabic outputs.

Despite being considered one of the largest Arabic open-sourced models that are pre-trained in massive datasets of 2.5 billion tokens, it has certain disadvantages that need to be considered when used for specific tasks. These include issues with "Out-Of-Vocabulary (OOV) words, which means the model cannot cover all Arabic words, particularly, rare or domain-specific words", the lack of domain knowledge, and the requirement for computational resources and memory.

EgyBERT is one of the Arabic large language models that was introduced specifically to address, analyse, and process the Egyptian dialect texts [24]. The model was pre-trained on a major dataset of Egyptian dialectal corpus, consisting of two novelty-developed corpora: Egyptian Tweets Corpus (ETC) of 34.3M tweets (2.5GB) and Egyptian Forums corpus(EFC) Online forum texts (7.9GB). [24] showed a developed model in dealing with the lack of dialect-specific models and compared the EgyBERT performance to the other five models on 10 Egyptian dialect tasks, such as sentiment analysis, gender identification, and sarcasm detection. The utilised BERT architecture with 12 layers and 75 heads. Advanced techniques were used to ensure dialect specificity and quality. The F1-score metric was used to evaluate the models' performance and accuracy. The model showed the highest average of the F1 score with 84.25% in different tasks and an accuracy of 87.33% compared to other existing models like MARBERTv2 and CamelBERT. Despite EgyBERT achieving significant results and filling an important gap in the Arabic NLP field and presenting tools and sources to enhance the process of Egyptian dialect texts, compared to other Arabic language models like MARBERT, which used datasets up to 128 GB, EgyBERT was pretrained on relatively small dataset 10.4 GB compared to other Arabic models like MARBERT with dataset 148GB, local machines with limited computational resources have been used to pre-train it. Additionally, the model lacks in dealing with other dialects and modern standard Arabic MSA, which affects its applicability in mixed dialects.

In general, there is no single model that fully captures all Arabic dialects. Despite models such as MARBERT, CamelBERT used subword tokenization to improve the models' ability to handle different dialects, but there are still performance variations across dialects, while the EgyBERT has customized tokenization to enhance the Egyptian Arabic dialect, superior to other models in this dialect, however, its generalization is restricted.

Previous PLM models, such as Arabic-BERT, employed pre-training techniques with dataset tokens ranging between millions to billions and were created with more specificity to the Arabic language structure. On the other hand, more advanced LLMs, including GPT-based models, used self-supervised learning methods across a variety of corpora and significantly bigger datasets with often trillions of tokens. This scale enables more generalization.

In addition to that, scaling up models has shown remarkable performance across various studies, such as those in [25], [26], [27]. For instance, the study of [25] investigated how scaling both data and models could affect model performance by compiling 529 GB pre-training corpus from different sources, such as common CAWL and Arabic dialects data and trained on (11-billion-parameter) in models called AraMUS. Researchers compared their performance with other less parameter models like SABER (with 369M parameters) and JABER (135 parameters). The AraMUS outperformed the other smaller models in multi-tasks and confirmed that increasing the model and dataset scale could improve models' performance. Alongside requiring significant computational resources, as 16 servers with 8 NVIDIA A100 GPU have been used for each training in this study.

Table 2. This table illustrates the most common Arabic pre-trained language models (PLMs) (early Arabic language models), including their timelines, preformed tasks, and limitations of each model.

Table 2. Most of the early Arabic language models (PLMs)

| No | Models Name | Year | Performed tasks (Results) | Limitation |
|----|-------------|------|---------------------------|------------|
| 1 | **ARENO** | 2021 | Improving Arabic NLP performance, particularly in handling rich morphology and dialectal variations, | Faced challenges with underrepresented dialects |

| | | | Improvements in tasks like text classification and sentiment analysis | |
|---|---|---|---|---|
| 2 | **Khaleej-BERT** | 2021 | Designed for Gulf Arabic dialects<br>Demonstrated superior accuracy in tasks specific to Gulf countries, such as dialect identification and sentiment analysis. | Its generalization to other Arabic dialects was limited |
| 3 | **Kuisai** | 2021 | Achieved reasonable performance improvements over baseline models. Its strength lay in handling Modern Standard Arabic (MSA) | Limited effectiveness with dialects |
| 4 | **DziriBERT** | 2021 | Focused on Algerian Arabic (Dziri dialect), improved dialect-specific tasks like text classification and sentiment analysis | Its performance in other Arabic dialects and MSA was not a focus |
| 5 | **CAMEL** | 2021 | Strong performance in MSA tasks such as named entity recognition (NER) and question answering. | Its ability to handle dialects was less explored |
| 6 | **AraT5** | 2021 | Based on the T5 architecture, excelled in text-to-text tasks such as summarization,<br>Question answering, translation<br>, and It set benchmarks for Arabic-specific NLP | Challenged with low-resource dialects |
| 7 | **AraELECTRA** | 2021 | Enhance efficiency and performance over BERT-based models in Arabic text classification and NER tasks.<br>Its pre-training objective (discriminating replaced tokens) enabled fast and more accurate learning | Struggling with some Arabic dialects<br>Required computational resources |
| 8 | **EgyBERT** [24] | 2024 | Superior performance in dialect-specific tasks such as sentiment analysis, text classification, and addressing a significant gap in dialectal NLP. | Cannot perform well on the Modern Standard Arabic (MSA) tasks |

## 4.2 Applications of Large Language Models to Arabic

Arabic large language models (LLMs) have the potential to transform various sectors in our social lives. For instance, LLMs could significantly contribute to preserving and revitalizing the rich cultural heritage of the Arabic language by accurately processing and producing content in both classical Arabic (CA) and Modern Standard Arabic (MSA). This would increase access to the native dialects on digital platforms and foster a deep connection to the language and its cultural background. In addition, Arabic LLMs can enhance education and economic development by being incorporated into educational and commercial tools, enabling learners and customers to interact with content using their preferred dialects.

It is also important to understand the potential for LLMs to inadvertently reinforce cultural bias. This highlights the need for diverse training data and culturally relevant benchmarks to avoid LLM bias, promote ethical AI use, and guarantee fair and accurate representation [28].

Therefore, several Arabic language models have been developed to serve different purposes. However, this paper will explain only the most common models in the NLP field according to their timelines, which selected starting with a seed subset of papers found by Google Scholar using the query "Large Language Models Arabic", and then we proceeded to explore the papers mentioned in this set of papers and the papers that cite them, and we tried to bring up LLMs research on various NLP applications.

### Arabic Text Translations

According to [11] This study examined how LLMs such as Bard, ChatGPT 3.5, and GPT4 could translate different Arabic dialects and Modern Standard Arabic MSA. The researchers used a dataset consisting of diverse Arabic dialects to test these Large models. Additionally, automated metrics like BLEU, COMET, and ChrF++ were used to evaluate the quality of text translation, along with human reviewers' evaluation to check whether the models correctly implemented the structure.

GPT4 is better in translation quality than Bard and Google Translate; in contrast, Bard was less accurate in following structures, and all models struggled when dealing with dialects that are less commonly spoken. Nonetheless, the study did not include all dialects, as the dataset was created manually, which limited the scope and restricted the evaluation.

### Arabic Grammatical Error Correction

Another study was conducted by [5], who explored the potential of LLMs performance, particularly ChatGPT, in Arabic Grammar Error Correction (AGEC). The study evaluated various prompting techniques such as Chain of Thought (CoT) and the experts' prompts to examine their effectiveness in enhancing LLMs in AGEC tasks. The findings showed that GPT-4, experts' prompts with few-shot learning, have presented a competitive performance on AGEC tasks in precise benchmarks and achieved an F1 score of 65.49. Promising results have been demonstrated in the potential performance of LLMs with grammar error correction tasks in low-resource languages, although further improvements need to match the fine-tuned models that are precisely trained on AGEC. Additionally, the use of the manual datasets limited the scope of the Arabic dialects and posed challenges for these models in handling complex Arabic features, such as orthography, morphology, and multi-dialects.

### Bilingual capabilities in both Arabic and English effectively

Bari et al. [29] introduced an extensive development of a large language model called ALLaM, which is tailored for language processing in both Arabic and English. The model was pre-trained on a mix of both languages' texts to improve proficiency in handling bilingual Arabic and English tasks. Parallel and translated data have been used to improve knowledge alignment among languages. The ALLaM model showed state-of-the-art results on different Arabic benchmarks as ACVA, MMLU, and Arabic Exams. it presented improvements in both languages in comparison to the baseline models.

On the other hand, training and optimizing LLM as ALLaM needs a significant computational resource, which may limit the academics' accessibility. In addition to that, the model efficiency across all different Arabic dialects still presents challenges due to the rich morphology and dialect variation of Arabic, which lead to ongoing and challenging training issues.

### Arabic AI Detectors

The performance of the GPTzero and OpenAI Text Classifier (AI detectors) on Arabic text was evaluated by [10]. The study introduced an Arabic dataset called AIRABIC, which consists of 500 passages, each containing both human-written text and AI-generating text (ChatGPT text) to assess the performance of these AI detectors on the Arabic dataset. The findings showed that GPTzero achieved an accuracy of 60% compared to the 50% for the OpenAI Classifier when analysing the Arabic human text. However, the accuracy of both detectors was notably reduced when detecting the diacritised Arabic human writing text. This finding highlights the importance of addressing these gaps when dealing with diacritical marks in the development of the detector process.

Within the same scope, [30]have introduced a new AI detector designed especially for the Arabic language using encoder-based transformer architecture. The study presented the utilization of two transformer-based models, which are trained on large datasets of both human-written text (HWT) and AI-generated text (AIGTs). GPT-3.5 was used to generate the Arabic AI text. The proposed models outperformed the current AI detectors in terms of recognising the Arabic HWTs and AIGTs. Additionally, a layer of dediacritization was integrated to improve detection accuracy by removing the diacritical marks. Nevertheless, this approach might not provide an exhaustive solution for improving AI detector proficiency, but it can be used as a first step toward more precise Arabic AI text detection. Furthermore, distinguishing between Arabic HWT and AI-generated text requires further exploration by researchers and developers, with the primary focus becoming more on text structure and style. Instead of considering diacritical marks as the main factor of differentiating between them. This is because ChatGPT and other LLMs do not use diacritics when generating AI text; recent processing techniques can be used to add diacritics if needed.

### Handling Linguistic Characteristics with Series of Arabic Models

According to the study [7]several versions of transformer-based models have been used, including ArabianGPT 0.1B and ArabianGPT 0.3B, which were specially developed to represent the distinctive linguistic characteristics of Arabic, such as its complex syntax and rich morphology. [7] demonstrated the significant performance of ArabianGPT models, especially with native Arabic text processing. For example, ArabianGPT 0.1B was fine-tuned and presented strong results in sentiment analysis, reaching 95% accuracy. The ArabianGPT 0.3B model was designed and fine-tuned specifically to deal with more complex NLP tasks such as summarizing and question-answering tasks, and it achieved a higher F1 score, reflecting better recall and precision. These models used the AraNizer tokenizer, which was designed especially to handle Arabic text by precisely tokening and processing its complex syntax and the rich morphological characteristics of the Arabic content. The tokenizer aids in efficiently training the models to better perform different Arabic language tasks. However, the capability for further improvement of ArabianGPT needs substantial computational resources for training the models; this may not be feasible for all researchers and academics. The models trained and focused on MSA are less effective in recognizing and capturing the linguistic nuances of the other Arabic dialects and limiting their utility with regional texts.

Table 3. This table shows the most advanced Arabic language models, including their timelines, developers, preformed tasks or results, and limitations of each model.

Table 3. Applications of large language models to Arabic language (LLMs)

| No | Models Name | Year | Developed by | Performed tasks (Results) | Limitation |
|---|---|---|---|---|---|
| 1 | **Habibi** | 2022 | Edinburgh University and Cambridge University | Achieved strong results in: Text Classification MSA-Focused Tasks. | Limited dialectal Performance |
| 2 | **KITAB** | 2022 | King Abdulaziz City for Science and Technology (KACST) | Text Classification Arabic Text Analytics Focus on structured & unstructured Arabic data processing | It may not handle some unstructured data types as effectively. |
| 3 | **Arabic-LLaMA** | 2022 | Meta AI Team and It Adapted From Meta's Llama | Outperformed general multilingual models in Arabic language tasks, particularly both zero-shot & few-shot scenarios | Handling diverse dialects remained a challenge |
| 4 | **NOOR** | 2022 | Abu Dhabi's Technology Innovation Institute (TII) | Improvements in translation quality between Arabic and other languages | Struggled with dialect-specific nuances |
| 5 | **Arabic-T5** | 2022 | The University Of California, Los Angeles (UCLA) | Strong performance in text generation and understanding tasks such as summarization and question answering. setting new benchmarks for Arabic NLP. | May not generalize well to all Arabic dialects |
| 6 | **Sahar** | 2022 | Saudi Company STC | Promising Results In Customer Service Applications, Enhancing Dialogue Systems And Conversational Agents In Arabic | Limited application beyond customer service tasks. |
| 7 | **Ruba** | 2022 | Emirates NBD | Focused On Financial Services, Like Document Classification And Customer Support Automation In Arabic | Limited to specific domains like finance |
| 8 | **AraPoemBERT** [31] | 2024 | Computer sciences department in Taibah University, Saudi Arabia | Focused on Arabic poetry analysis and generation, capturing the stylistic and structural nuances of Arabic poetic forms, showcasing state-of-the-art performance in poetry-specific NLP tasks. | Limited to poetry and may not generalize to other types of text. |
| 9 | **ALLaM** | 2024 | Saudi Data and Artificial Intelligence Authority (SDAIA) with IBM | Improved proficiency in handling bilingual Arabic and English tasks | Needs a significant computational resource |

# 5. Arabic benchmark and datasets

In order to facilitate pre-training large language models with high-quality corpora, several benchmarks, and datasets have been built for training and evaluating various natural language processing tasks (NLP), including machine learning translation, question answering (QA), named entity recognition (NER), text classification, dialect classification, and sentiment analysis. The next paragraphs will demonstrate a list of commonly used Arabic benchmarks and datasets in NLP tasks, starting with general NLP benchmarks and then specific NLP datasets.

## 5.1 General NLP benchmarks

**Arabic language understanding evaluation (ALUE) benchmark**: It is considered as one of the widely recognized benchmarks based on general language understanding evaluation (GLUE), which evaluates Arabic NLP models on a variety of language tools, especially in academic institutions use MSA content, and other NLP tasks such as NER, sequence labeling, regression, paraphrasing, classification, and sentiment analysis. However, the ALUE generally concentrates on Modern Standard Arabic (MSA), which limits its applicability to the Arabic dialects, which widely used in the real world. Furthermore, some tasks could not have enough domain variety, which might cause overfitting on benchmark patterns.

**ArabicGPT benchmark:** This benchmark was created to assess Arabic GPT-based models in terms of the following tasks: summarization, text generation, and dialogue interpretation understanding. It can ideally be used with generative model applications such as content creation and chatbots. Whereas this benchmark frequently relies on synthesized datasets, which can produce biases and could not accurately represent the variety of domains and dialects, it might not effectively express conversational details, especially with informal language or limited resources.

**Nuanced Arabic dialect identification (NADI) benchmark:** The main focus is on Arabic dialect identification at both country and regional levels and categorizes dialects into groups such as Gulf, Dziri, Egyptian, Moghebi, etc. The NADI benchmark is primarily dependent on the representation and quality of the social media content, which could not be adapted to a formal context. Therefore, it's commonly used in voice assistance and region content applications, and sociolinguistic studies, where accurate dialects are crucial.

## 5.2 Specific NLP Datasets

**Arabic sentiment tweets dataset (ASTD):** An Arabic tweets dataset with sentiment labels like positive, negative, and neutral with about 10000 tweets in size.

**Large Arabic Book Reviews (LABR):** a collection of Arabic book reviews with sentiment scores, about the size of 63000 reviews.

**SemEval 2017 tasks 4 subtasks:** special Arabic sentiment analysis tweets dataset Sentiment

**Arabic-named entity recognition corpus (ANERCrop):** Arabic annotated corpus containing categories such as locations, organization, and persons.

**Automatic content extracting (ACE):** As part of this multilingual dataset, Arabic text is covered with named entity and relation annotations.

**Arabic SQUAD:** This dataset has been modified for Arabic with an emphasis on questions answered and reading comprehension.

**TYdi QA:** This is a multilingual dataset that answers questions, including a considerable amount of Arabic content.

**OSCAT4 Shared Tasks**: The main focus of this Arabic dataset is on detecting offensive languages in tweets, working with a binary classification such as offensive/non-offensive.

**ATD (Arabic Text Dataset):** Sports, politics, and technology are just a few categories covered by this text classification dataset.

**Parallel Arabic Dialect Corpus (PADIC):** The dataset contains a corpus for dialect classification and machine translation.

**AraT5 Dataset:** text-to-text, such as summarisation and translation, and QA.

**Multi-Arabic dialect applications and resources (MADAR):** The dataset contains a corpus of 25 different dialects.

**Arabic English parallel Croups (Ar-En):** Arabic to English machine translation.

**Qatar Arabic Language Bank (QALB):** Arabic grammar error correction dataset.

**OpenITI**: useful for Pre-training models on modern Arabic text.

Tashkeela Dataset: diacritization dataset with a size of 75 million words.

**Wikipedia Arabic:** collection of Arabic Wikipedia articles dataset for pre-training and other purposes.

Figure 2 shows the categories of each dataset based on their role in real life. For instance, ATD and OSACT4 datasets are commonly used in text classification tasks, while PADIC and MADAR are dialectal datasets, etc. Whereas other datasets such as QALB and Tashkeela are employed specifically for unique Arabic language structure tasks such as grammar error correction and Tashkeela (diacritic) marks.

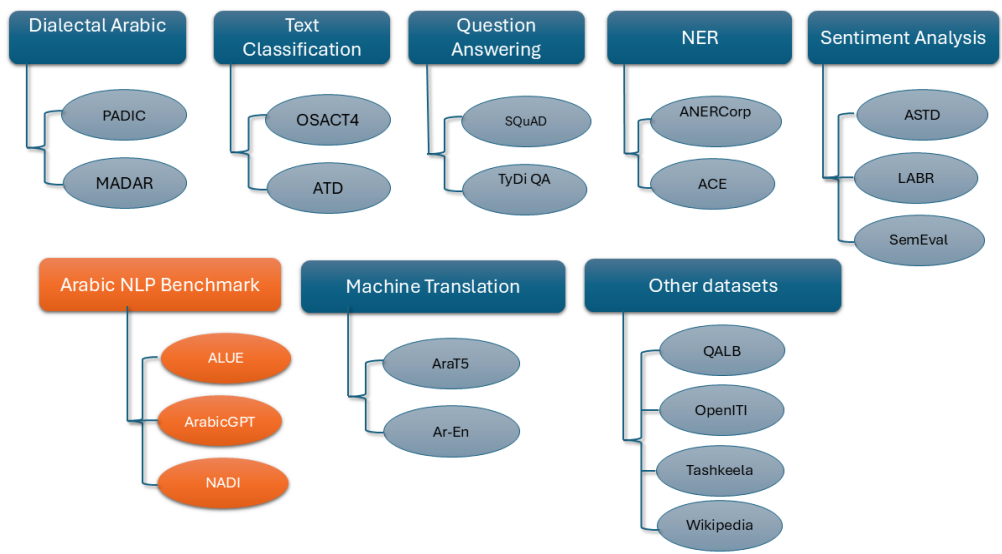

**Fig. 2.** Arabic dataset categories

## 6. Trends and Observations

### 6.1 Rapid development in Arabic Language models.

The last three years have shown important progress for specific tasks in large language models, such as ArabianLLM, AraGPT, ALLaM, Arabic-LLaMA, etc. These models and others reflect the increasing trend to adjust state-of-the-art LLMs to the unique features of Arabic language linguistics, for example, its Orthography, rich morphology, complexity, and diacritical marks. etc. Academics and researchers are continuously employing techniques of fine-tuning and transfer learning to improve these models' performance, especially in Arabic NLP tasks ranging from machine translation, text classification, grammar correction, and others.

### 6.2     Covering the Arabic NLP Resource Gap

A clear and notable observation is increasing efforts in improving and creating high-quality corpora for the Arabic language. Initiatives such as Wikipedia datasets, OpenITI [32], name entity recognition, and annotated corpora in sentiment analysis contribute to dealing with the historical scarcity of Arabic NLP sources. This trend shows the significance of Arabic-specific datasets in enhancing the robustness and the ability to generalize trained LLMs for the Arabic content and contributes to achieving significant results in filling and decreasing an important gap in the Arabic NLP field.

### 6.3     Growing Focus on Arabic Dialects

Another trend is a clear focus in the direction of processing Arabic dialects, which are commonly and widely spoken in different Arabic countries but insufficiently represented in the Arabic written content.  More models are being developed specifically to address tasks such as distinguishing between dialects, performing sentiment analysis, and enabling conversational AI for Arabic regional dialects [33] .

## 7. Conclusions and Future Work

This paper provides an overview of large language models (LLMs) applications to Arabic contexts and reviews related studies, with focus on recent large language models such as ChatGPT. It highlighted the challenges of the Arabic language, including the rich morphology, orthography, diacritical marks, ambiguity, and diverse dialects. These challenges have contributed to a relatively limited number of Arabic studies compared to other languages such as English. The paper also briefly reviews related Pre-trained Language Models (PLMs), including AraBERT, MARBERT, and QARIB to outline their development and timeline. It presents also the development of the recent Arabic language models and common Arabic benchmark datasets, which are designed to serve various purposes and tasks, including Arabic AI detectors, dialects models, and Grammar error correction models. It demonstrates the growing attention of researchers and students given to Arabic NLP tasks recently and shows how such effort could pave the way for the development of additional Arabic models, tools, and resources to address other important tasks such as text simplification and further narrow the gap between Arabic NLP and work done on other languages.

### 7.1     Future Directions

This section proposes recommendations and outlines potential paths for further research, addressing current limitations and exploring emerging questions.
1. Future studies should concentrate on training and enhancing diverse Arabic datasets for both MSA and different dialects, as well, improving LLMs performance by developing various open-access corpora, including different domains such as news, social media, and literature. etc.
2. Tackling the challenges of diverse dialects by exploring effective methods to integrate dialectal variations into LLMs and leveraging transfer learning technologies to aid models in generalisation across dialects would be valuable.

3. Proposing new metrics to evaluate the models' performance and reflect all the nuances of the Arabic language.
4. Research should examine how biases in Arabic datasets may impact models' outputs by developing appropriate cultural criteria to ensure fairness in the Arabic NLP field.
5. Further study is needed to determine how LLMs can be adapted for practical use in real-world applications of the Arabic-speaking world, including media, education, and government services, as this will offer important insights.
6. Researchers should look at hybrid model workflows that could support human content producers rather than replace them, enabling a cooperative strategy between AI and human knowledge.